# How to enhance learning of robotic surgery gestures? A tactile cue saliency investigation for 3D hand guidance

Gustavo D. Gil[1], Julie M. Walker[2], Nabil Zemiti[1], Allison M. Okamura[2], Philippe Poignet[1]

[1] LIRMM, University of Montpellier, CNRS, Montpellier, France
[2] Department of Mechanical Engineering, Stanford University, Stanford, CA 94305, USA

gustavo.gil@lirmm.fr

## INTRODUCTION

The current generation of surgeons requires extensive training in teleoperation to develop specific dexterous skills, which are independent of medical knowledge. Training curricula progress from manipulation tasks to simulated surgical tasks [1]–[3] but are limited in time due to the need for proper mentoring for each trainee.

We propose to integrate surgical robotic training together with Haptic Feedback (HF), as is illustrated in Fig. 1. The reason is that a good use of HF during training can improve skill acquisition [4]–[7].

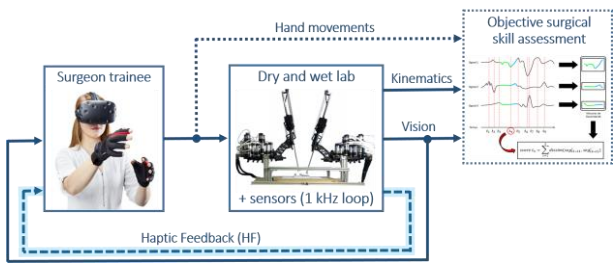

**Figure 1.** Full dry and wet lab approach (non-clinical setup) of our complete robotic surgery training system. During the execution of the surgical gestures the highlighted haptic feedback loop transmit to the surgeon trainee timely information about his/her performance.

In this paper we introduce an exploratory work on a portable haptic device designed to offer hand guidance in 3D space. The idea is that the HF can be felt by the trainee "as if" a force was directing his/her entire hand through of the surgical gesture execution.

Mainly, we create HF signals by stretching the skin of a pair of fingers (e.g., Thumb-Index). Our haptic device acts on the *finger pads* of the *fingertips* (see Fig. 2), in order to stimulate the cutaneous mechanoreceptors.

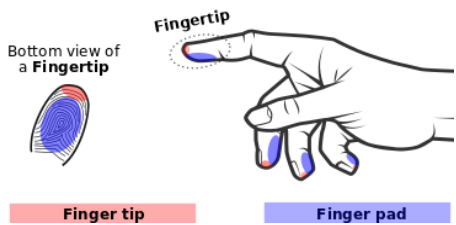

**Figure 2.** The colored areas indicate the jargon for different zones of the human fingertips (i.e., finger tip and finger pad).

## MATERIALS AND METHODS

The device consists of a pair of servomotors mounted in a 3D printed handle, as shown in Fig. 3. Each motor rotates a lever arm to stimulate the trainee's finger pads of the thumb and index distal phalanges, while the trainee holds it in his/her hand.

The stimulations, called *haptic signals*, consist of to stretch the skin on the finger pads by rotating through a semicircular arc (see schema in Fig. 3). Thus, different movements of the arms give to the user the sensation that their hand is being pulled in certain directions. We called these senstations *haptic cues*.

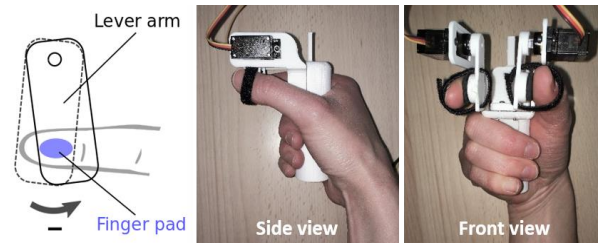

**Figure 3.** Scheme for one actuator of the haptic device (working principle) and views of the entire device.

The experiment presented here aims to identify two key aspects: (1) cue saliencies or cue clarity; and (2) if there is a common perception of tactile cues between users. The first aspect is related to stimulus-cue repeatability by the users, while the second one is related to the identification of a common stimulus-cue set among different users.

We investigate the feasibility to induce three types of hand movements, or 3 Degrees of Freedom (DoF). Namely: right/left wrist twisting; hand moving forward/backward; and hand tilting up/down.

The experiment involved 8 right-handed engineers. 24 different stimuli were applied in a pseudorandom sequence of 240 stimuli per trial (each stimulus 10 times). Trials included two time breaks in order to release the device, relax the hand, and continue the experiment.

For each *haptic signal* (i.e., stimulation induced), users reported the cue felt from six empirically predetermined options: (1-2) hand moving Forward/Backward; (3-4) Tilt Up/Down; and (5-6) Twist Left/Right.

During the trial, the users trigger each haptic stimulus by using a key of a computer keyboard. Then, the cue felt was reported by a "number key" assigned.

Sumarizing the test method, we executed a system identification procedure to investigate the responses of the tactile sense of each user. Our aim was to identify stimuli-cues that can be used as commands to guide the user's hand in 6 different directions (i.e., in 3 DoF).

## RESULTS

Each trial took about 45 minutes plus extra time for the familiarization with the procedure. People were constrained to select between 6 possible directions. It was noticeable that most of the time participants chose a direction after triggering each stimulus just 1 or 2 times. The multidimensional nature of the results is expressed by colored marks in a two-dimensional map (Fig. 4). The meaning of the features of this map are:

- the colors encode the type of cue most often selected by the users (e.g., hand pulled Forward/Backward, hand Tilt Up/Down, and hand Twist Left/ Right).
- the x-y location of each mark corresponds to a combination of angular displacement for each servomotor (stimulus related).
- the mark size represents the cue saliency or cue clarity. The bigger diamonds correspond to the most salient cues.

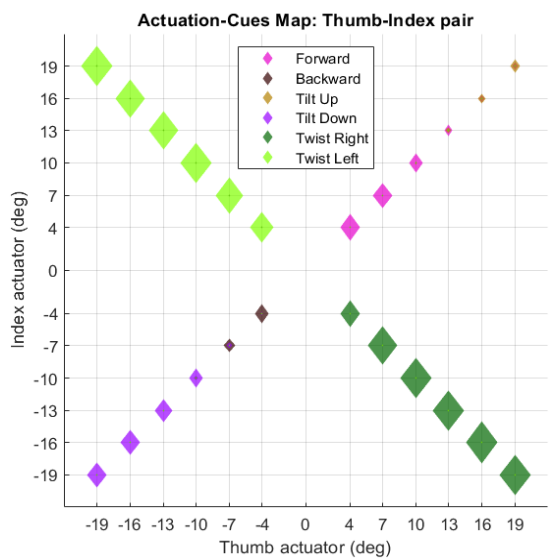

**Figure 4.** Cues identified by 8 users in the exploratory experiment. The mark size relates to cue saliency, the color to cue type, and the location to stimulus direction and magnitude (servomotor actuation).

From the data obtained in this exploratory study, we can agree that the device is able to induce at least two clear tactile cues, called hand Twist Left and hand Twist Right. These two cues are very salient for several combinations of servomotor values (see the 2$^{nd}$ and 4$^{th}$ quadrants of the Actuation-Cues Map), identified more than 85% of the time.

Conversely, the cues labeled as hand Tilt Up and hand moves Backward, are not as salient but are present. For the remaining cues, hand moves Forward and hand Tilt Down, it seems that a trend appears, then this haptic device deserves more research.

## DISCUSSION

Interestingly, our haptic device is able to induce a directional cue (i.e., a feeling of direction) without prior training. Thus, the principle of skin stretch [8], [9] explored here allows us to envisage ungrounded and low-cost haptic devices for hand guidance.

In the experiment some people had problems with the thumb, which occasionally slipped out from the haptic device. This issue can be related to the symmetrical arrangement of the device. A modified device with a better ergonomics is foreseen for future test campaigns. A good alternative is proposed in [10] using 4 motors.

To conclude, we see the Haptic Feedback (HF) as a key ingredient of our recipe for surgical robotic training. There is potential for HF to speed training times for surgeon residents, while increasing development of relevant visual-motor skills. Our ongoing work is related to integrate the HF in the RAVEN II platform.